\documentclass[letterpaper, 10 pt, conference]{ieeeconf}  
\IEEEoverridecommandlockouts

\usepackage{cite}
\usepackage{amsmath,amssymb,amsfonts}
\usepackage{algorithmic}
\usepackage{graphicx}
\usepackage{textcomp}
\usepackage{xcolor}
\usepackage{hyperref}

\usepackage{multirow}
\usepackage{dcolumn}
\usepackage{diagbox}

\usepackage{color,soul} 
\definecolor{red}{rgb}{1.00,0.00,0.00}
\definecolor{blue}{rgb}{0.00,0.00,1.00}
\definecolor{green}{rgb}{0.30, 0.50,0.00}

\newcommand{\cblue}[1] {\textcolor{blue}{#1}}

\usepackage{multirow}

\def\BibTeX{{\rm B\kern-.05em{\sc i\kern-.025em b}\kern-.08em
    T\kern-.1667em\lower.7ex\hbox{E}\kern-.125emX}}
    
\begin{document}



\title{Enhancing Fine-Grained 3D Object Recognition \\using Hybrid Multi-Modal Vision Transformer-CNN Models}

\author{Songsong Xiong$^{1}$, Georgios Tziafas$^{1}$ and Hamidreza Kasaei$^{1}$
\thanks{$^{1}$Department of Artificial Intelligence,
        University of Groningen, Groningen, The Netherlands\newline 
        {\tt\small \{s.xiong, g.t.tziafas,hamidreza.kasaei\}@rug.nl}}}

\maketitle
\thispagestyle{empty}
\pagestyle{empty}

\begin{abstract}
Robots operating in human-centered environments, such as retail stores, restaurants, and households, are often required to distinguish between similar objects in different contexts with a high degree of accuracy. However, fine-grained object recognition remains a challenge in robotics due to the high intra-category and low inter-category dissimilarities. In addition, the limited number of fine-grained 3D datasets poses a significant problem in addressing this issue effectively. In this paper, we propose a hybrid multi-modal Vision Transformer (ViT) and Convolutional Neural Networks (CNN) approach to improve the performance of fine-grained visual classification (FGVC). To address the shortage of FGVC 3D datasets, we generated two synthetic datasets. The first dataset consists of $20$ categories related to restaurants with a total of $100$ instances, while the second dataset contains $120$ shoe instances. Our approach was evaluated on both datasets, and the results indicate that it outperforms both CNN-only and ViT-only baselines, achieving a recognition accuracy of $94.50\%$ and $93.51\%$ on the restaurant and shoe datasets, respectively. Additionally, we have made our FGVC RGB-D datasets available to the research community to enable further experimentation and advancement. Furthermore, we  successfully integrated our proposed method with a robot framework and demonstrated its potential as a fine-grained perception tool in both simulated and real-world robotic scenarios.

\end{abstract}

\section{Introduction}

As society continues to grow, labor shortages have become increasingly prevalent. Consequently, robots are gaining popularity in human-centered environments~\cite{INTR03,INTR04}. To safely operate in such domains, the robot should be able to recognize fine-grained objects accurately. For example, a restaurant robot must be able to categorize drinks with similar packaging but varying attributes. Similarly, a service robot in the shoe store is required to sort shoes with comparable appearances. Fine-grained visual categorization (FGVC) has recently received considerable attention~\cite{INTR06} as it aims to identify sub-categories within the same basic-level classes. However, it remains a challenging task due to the high intra-category and low inter-category dissimilarity issues~\cite{INTR05}. Furthermore, the performance of FGVC is often hindered due to limited available datasets~\cite{INTR06}.

Recently, the majority of studies have been focused on RGB fine-grained recognition. These studies include the CUB-200-2011 dataset~\cite{INTR06}, Oxford Flowers dataset~\cite{INTR07}, Aircraft dataset~\cite{INTR08}, and Pets dataset~\cite{INTR09}. Additionally, some studies have employed both RGB and depth sensors to perform a variety of robotic tasks, including object classification~\cite{INTR10, INTR11, INTR12,INTR14, INTR15} and action recognition~\cite{INTR13}. These studies have demonstrated that using multi-modal object representations enhances recognition accuracy.

To the best of our knowledge, there are only two FGVC RGB-D datasets currently available. The first dataset is limited to hand-grasp classification, while the second dataset is centered around vegetables and fruits, but unfortunately, it is not publicly available~\cite{INTR16, INTR17}. Furthermore, due to their small scale, these datasets impose restrictions on the performance of deep learning methods for FGVC~\cite{INTR19}. Consequently, there is a lack of large-scale RGB-D datasets that can be utilized for FGVC purposes. To address the shortage of FGVC 3D datasets, we generated two synthetic datasets. The first dataset consists of $20$ categories related to restaurants with a total of $100$ instances, while the second dataset contains $120$ shoe instances.

\begin{figure}[!t]
\vspace{1.5mm}
\centerline{\includegraphics[width=\linewidth]{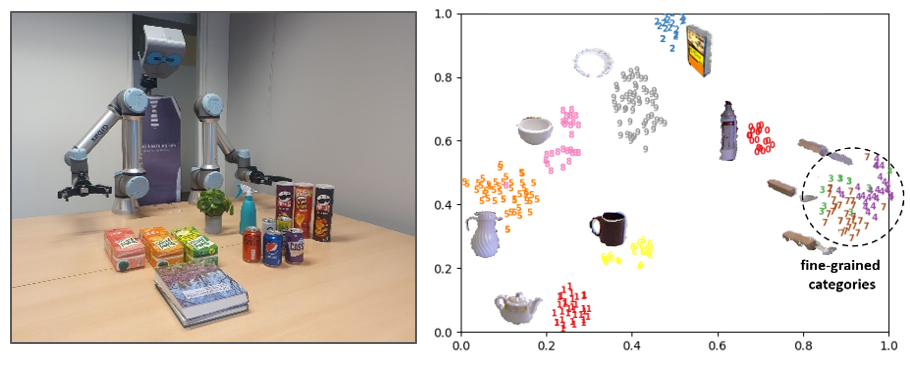}}
\caption{An example of fine-grained object visual classification (FGVC) scenario: (\textit{left}) The robot leverages an RGB-D camera to perceive the environment ; (\textit{right}) The distribution of various object categories across the RGB-D feature space is displayed using a t-SNE plot. This plot reveals that distinguishing between fine-grained categories like knife, fork, and spoon is more challenging than distinguishing between basic-level categories like mugs and bottles.}
\label{robot1}
\end{figure}

\begin{figure*}[!t]
\includegraphics[width=\linewidth]{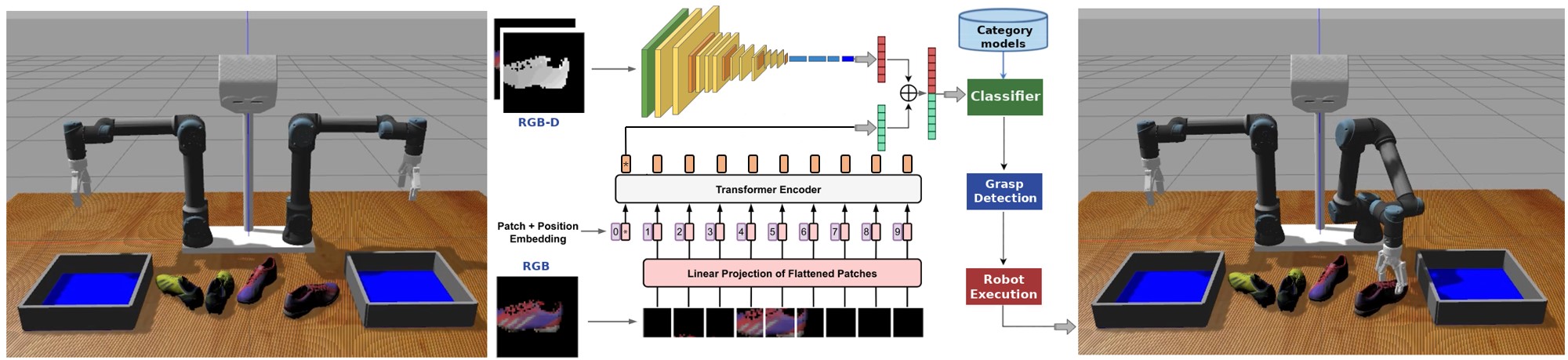}
\caption{A dual-arm service robot is used to successfully sort fine-grained shoe objects:
(\textit{left}) To accomplish this task successfully, the robot should recognize all fine-grained shoe objects first and then place similar shoes in the same basket (\textit{shoes}); (\textit{center}) the proposed hybrid multi-modal approach receives RGB-D images of each object and passes them through a CNN network and a  ViT simultaneously. The CNN network is responsible for capturing local information of the object, while the ViT is used to encode global features of the objects. The obtained representations are then fused and used for object recognition purposes; (\textit{right}) Afterward, the robot detects a proper grasp configuration for the target object \cite{kasaei2023mvgrasp}, and then grasps and manipulates the object into the basket.}
\label{fgvc_overview}
\end{figure*}

Furthermore, we propose a hybrid multi-modal approach based on ensembles of CNN-ViT networks to enhance the accuracy of fine-grained 3D object recognition. An overview of our approach is shown in Fig.~\ref{fgvc_overview}. 
We performed extensive sets of experiments to assess the performance of the proposed approach. Experimental results show that our multi-modal approach surpasses corresponding unimodal CNN-only and ViT-only approaches in recognition accuracy. Additionally, we have successfully integrated our proposed method with a robot framework and demonstrated its potential as a fine-grained perception tool in both simulated and real-world robotic scenarios. 

In summary, our key contributions are twofold:

\begin{itemize}

\item We propose a hybrid multi-modal approach based on ViT-CNN networks to enhance fine-grained 3D object recognition. 
\item To the best of our knowledge, we are the first group to build the publicly available 3D object datasets for fine-grained object classification. The datasets are publicly available online at:  
\textit{\cblue{\href{https://github.com/github-songsong/Fine-grained-Pointcloud-Object-Dataset}{https://github.com/github-songsong/Fine-grained-Pointcloud-Object-Dataset}}}

\end{itemize}

\section{RELATED WORK}


Various studies have been carried out on fine-grained object recognition, which can be classified into three categories based on their approach, namely localization methods, feature encoding methods, and transformer methods, as discussed in~\cite{INTR05}. 

\noindent \textbf{Localization-based FGVC methods}:
These methods focus on identifying discriminative partition areas between instances by training a detection model, and then classifying using the trained model. For examples, Branson et al.\cite{RELA11} and Wei et al.\cite{RELA12} proposed supervised learning of localization via part annotations. However, due to the high cost and limited availability of part annotations, weakly supervised learning using image labels alone has gained more attention. Yang et al.~\cite{RELA13} introduced a re-ranking method to enhance region representations for global categorization. Unlike our approach, these methods require specially designed models to identify potential areas, and the selected sections must still undergo classification via a backbone model.

\noindent \textbf{Feature-encoding methods:} These methods aim to enhance the object representation to achieve better classification performance. Our approach falls into this category. Yu et al.~\cite{RELA14} enhanced the object representation by utilizing the hierarchical bilinear pooling function, which combines the multiple cross-layer bilinear features. Zheng et al.~\cite{RELA15} proposed 
a deep bilinear transformation block, which can be deeply stacked in convolutional neural networks to learn fine-grained image representations. In particular, they uniformly categorized the input channels into several semantic groups and then generated a compact representation for FGVC. As these methods use a single encoder and only RGB data, their performances are limited~\cite{RELA16}. To address these limitations, we consider multi-modal and ensemble of ViT and CNN models to handle the FGVC task.

\begin{figure*}[!t]
\vspace{2mm}
\begin{tabular}{ccc}
    \includegraphics[width=0.31\linewidth,height=0.2\textwidth]{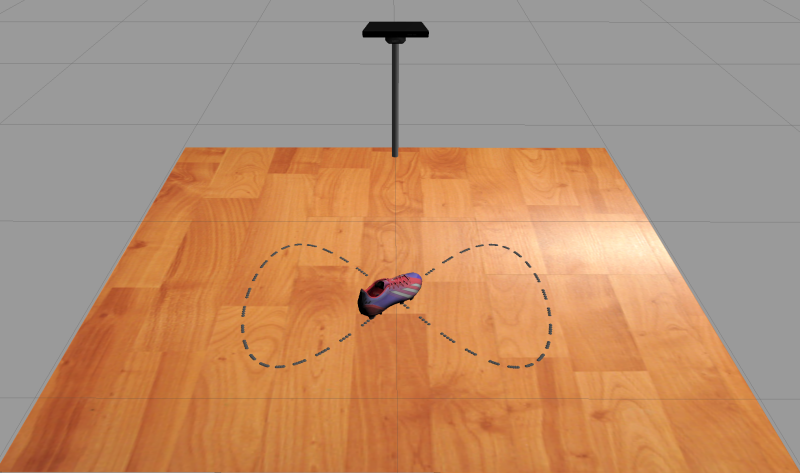} & 
    \includegraphics[width=0.31\linewidth,height=0.2\textwidth]{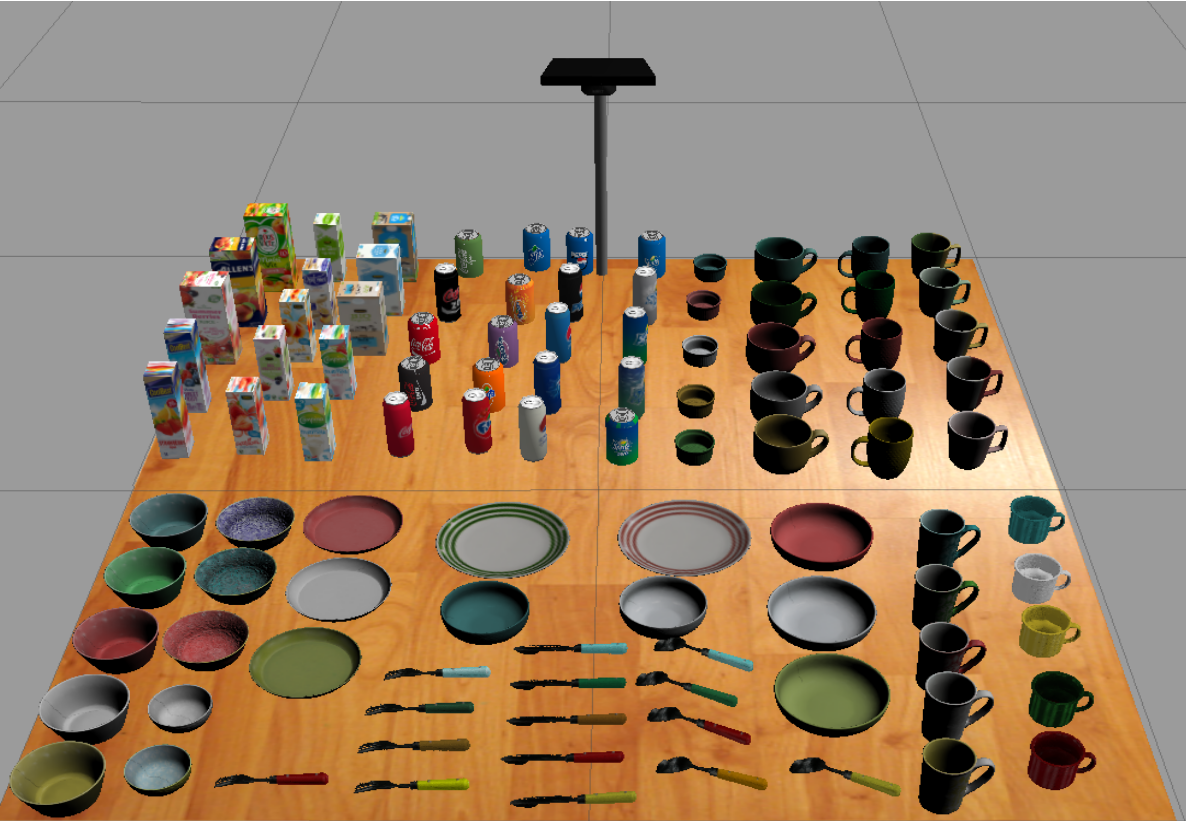}  & \includegraphics[width=0.31\linewidth,height=0.2\textwidth, trim= 0mm 0mm 0mm 0mm, clip = true ]{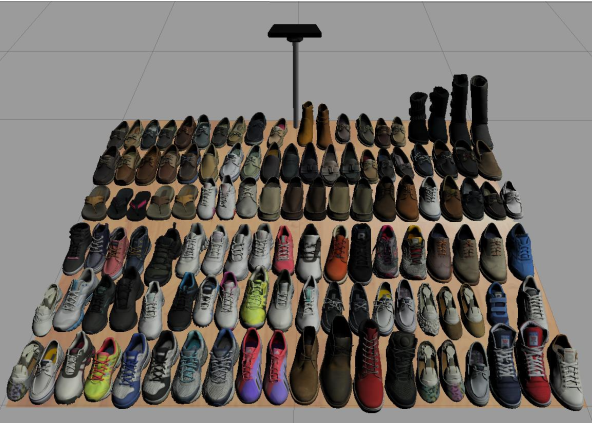}
\end{tabular}
\caption{Generating fine-grained point cloud object datasets in Gazebo environment: (\textit{left}) Our setup consists of a table and a Kinect camera mounted on a tripod. We place an object on the table and move it along a fixed trajectory in front of the camera. We segment and record the point cloud of the object in each step; (\textit{center}) Fine-grained restaurant objects used to record the dataset; (\textit{right}) fine-grained shoe objects used to record the dataset. 
}
\label{fgvc_dataset}
\end{figure*}

\noindent 
\textbf{Transformer methods:} In recent years, transformers have shown remarkable progress in Natural Language Processing~\cite{RELA17,RELA18}. As a result, more studies have applied transformers to computer vision tasks, including object detection~\cite{RELA19,RELA20}, segmentation~\cite{RELA21,RELA22}, and object tracking~\cite{RELA23}. In particular, using ViT models for FGVC has gained increasing popularity~\cite{RELA27,RELA28}. For instance, Dosovitskiy et al.\cite{RELA24} proposed the ViT model, which demonstrated superior performance in image classification. Subsequently, Swin\cite{RELA25}, DeiT~\cite{RELA26}, and MAE~\cite{comp03} were introduced for various computer vision tasks. He et al.~\cite{RELA16} extended the ViT-only model to FGVC and evaluated the proposed approach on traditional RGB-only datasets (e.g., various bird species, and car models).

Many researchers have recently utilized CNN-only or ViT-only models to tackle FGVC with the RGB-only datasets.
Ullrich et al.~\cite{RELA28} leveraged a multi-CNN network to extract RGB and Depth images for 3D object recognition. 
The RGB and Depth image representations from CNN-only models separately are also used for the single-view FGVC~\cite{INTR17}. 
The performance spectrum of CNNs and ViTs, however, is confined under their fixed architectures. With the training dataset increasing, their accuracy can be improved but only approach the maximum of their fixed capacity. 
To improve the performance of single-view FGVC, we proposed the hybrid multi-modal approach based on ViT and  CNN networks. 


\noindent\textbf{Fine-grained object datasets:} In recent studies on FGVC, Nilsback~\cite{INTR07} contributed a fine-grained flower dataset with 17 different species for FGVC, followed by the fine-grained Birds dataset containing 11788 images from 200 bird species~\cite{INTR06}. 
Since then, FGVC has gradually gained more attention. For example, the Standford Dogs~\cite{INTR09} and Cars~\cite{RELA01} datasets for FGVC were published, respectively. Fine-grained VegFru~\cite{RELA02} consisting of vegetables and fruits, and Kuzushiji-MNIST~\cite{RELA03}, have been introduced recently.

Considering the limitation of the RGB-only data, RGB-D images rapidly emerged in computer vision tasks due to providing additional rich information.
For example, Andreas et al.~\cite{ RELA04} released an object segmentation dataset, which comprises of 111 RGB-D images with stacked and occluding objects on the table. 
The UR Fall detection dataset, contributing to tracking human skeletal research, was created by Bogdan et al~\cite{RELA05}. 
Several datasets for basic-level object recognition tasks also released, such as the Wahington RGB-D object dataset with 300 instances of household objects~\cite{RELA06}, NUY Depth V2~\cite{RELA07}, and Restaurant object dataset ~\cite{RELA08}. 
However, with the exception of the hand-grasp dataset~\cite{INTR16} and the RGB-D dataset with vegetables and fruits~\cite{RELA09} (which is not publicly available), almost all publicly available FGVC datasets consist of RGB-only images or grayscale-only images. To address the lack of RGB-D dataset for FGVC, we created two FGVC RGB-D datasets consisting of 120 categories of shoe objects (including 7200 instance views) and 20 categories of restaurant objects (with 1200 instance views).

\section{METHOD}

As shown in Fig.~\ref{fgvc_overview}, we use RGB-D data as inputs for our model.
The model then processes the inputs and generates the corresponding fine-grained representation that will be utilized to learn and recognize objects. In this section, we first introduce the process of generating fine-grained RGB-D object datasets, followed by discussing our hybrid multi-modal CNN-ViT networks.


\subsection{Generating fine-grained object datasets}

One of the primary goals of this research is to generate RGB-D datasets for FGVC tasks. Towards this goal, we develop a simulation environment in Gazebo to record data, which consists of a table and a Kinect camera mounted on a tripod (see Fig.~\ref{fgvc_dataset}). We import the 3D model of several restaurant fine-grained objects and shoe fine-grained objects from publicly available resources and google objects~\cite{INTR18}. 
In our fine-grained RGB-D object datasets, there are 1200 restaurant object views categorized into 20 categories (\textit{100 objects}) and 7200 shoe views organized into 120 cataergories.

As depicted in Fig.~\ref{fgvc_dataset} (\textit{left}), we placed each object on top of a table and moved it in front of the camera along a predetermined trajectory, capturing 60 partial views of each shoe instance and 12 partial views of each restaurant instance. Our datasets present a highly challenging scenario for FGVC tasks, as the objects included have nearly identical properties in terms of geometry, textures, and colors, making it difficult even for humans to distinguish between them (see Fig.~\ref{fgvc_dataset} \textit{center} and \textit{right}).

\subsection{Multi-modal representations with ViT and CNN}
In recent research on deep learning approaches for FGVC, insights include that CNNs often outperform ViT on small-scale data, while ViT can gradually outperform the CNN as data size increases ~\cite{RELA24}. As a result of differences in representation characteristics, CNN and ViT can focus on learning local and global information of objects, respectively~\cite{Method04}.
To capture more comprehensive features of an object, we propose and compare two hybrid multi-modal approaches that takes into account both ViT and CNN encoders.


To encode an object for fine-grained classification, we use the orthographic projection method to extract RGB and Depth images from the object's point cloud, as described in~\cite{kasaei2020orthographicnet}. These images are then inputted into both a ViT and a CNN encoder, which have been pre-trained on ImageNet. The resulting representations from the RGB and depth images are fused to create a multi-modal representation for the given object.
In the first approach, we use a ViT network to encode the RGB image and a CNN architecture to encode depth data. In the second method, we consider another CNN to encode RGB image too. Afterward, all the obtained representations from ViT (RGB) and CNN (RGB-D) will be fused using an element-wise pooling function (e.g., average, maximum) or appending operator to form a descriptive feature vector for the given object. It is important to note that if the length of feature vectors obtained from different networks is not the same, we concatenate the feature vectors to form a unified representation for the object.
We then use various classifiers to evaluate the classification performance. Our findings reveal that the K-nearest neighbor (kNN) classifier outperforms other classifiers, particularly with limited training data. As a result, we select the kNN classifier with the Motyka distance function to measure the similarity between objects. Through multiple experiments, we have found that the value of k is a crucial parameter for the kNN. Our findings suggest that setting k to 1 produces the best results. An overview of our hybrid multi-model approach is shown in Fig.~\ref{fgvc_overview}.

\section{Result and Discussion}

We conducted several experiments to evaluate the performance of the proposed method using fine-grained object datasets. To select the best classifier, we first evaluate the performance of various classifiers using the restaurant fine-grained object dataset. 
We then performed an extensive set of experiments in both datasets using the 10-fold cross-validation algorithm~\cite{Method02}. In particular, we randomly divided the dataset into ten folds, one fold serving as test data and the remaining nine serving as training data in each iteration. This experiment is repeated ten times, so each fold is used as test data once. To measure the performance of object recognition we used instance accuracy ($\frac{\#\operatorname{true~predictions}}{\#\operatorname{predictions}})$.

\subsection{Classifier comparison}

\begin{figure}[!h]
    \begin{tabular}{cc}
        \hspace{-0.35cm} \includegraphics[width=0.5\columnwidth]{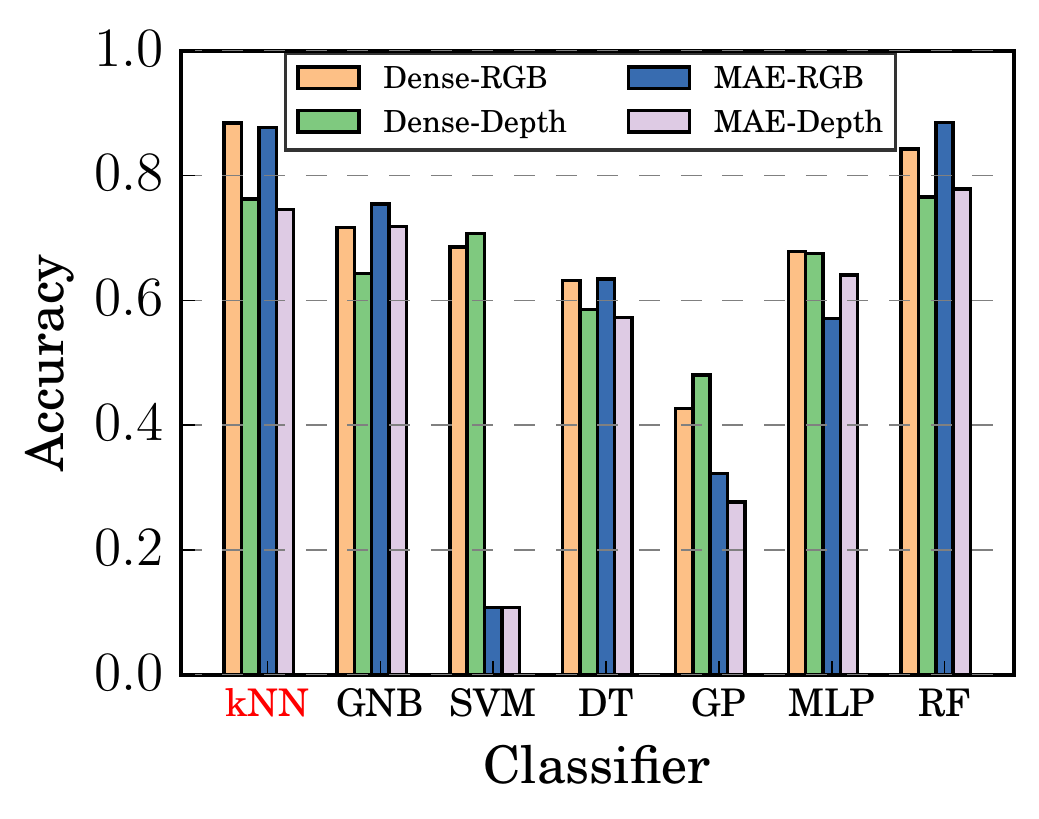} & \hspace{-0.65cm}
        \includegraphics[width=0.5\columnwidth, trim= 0mm 0mm 0mm 0mm, clip = true]{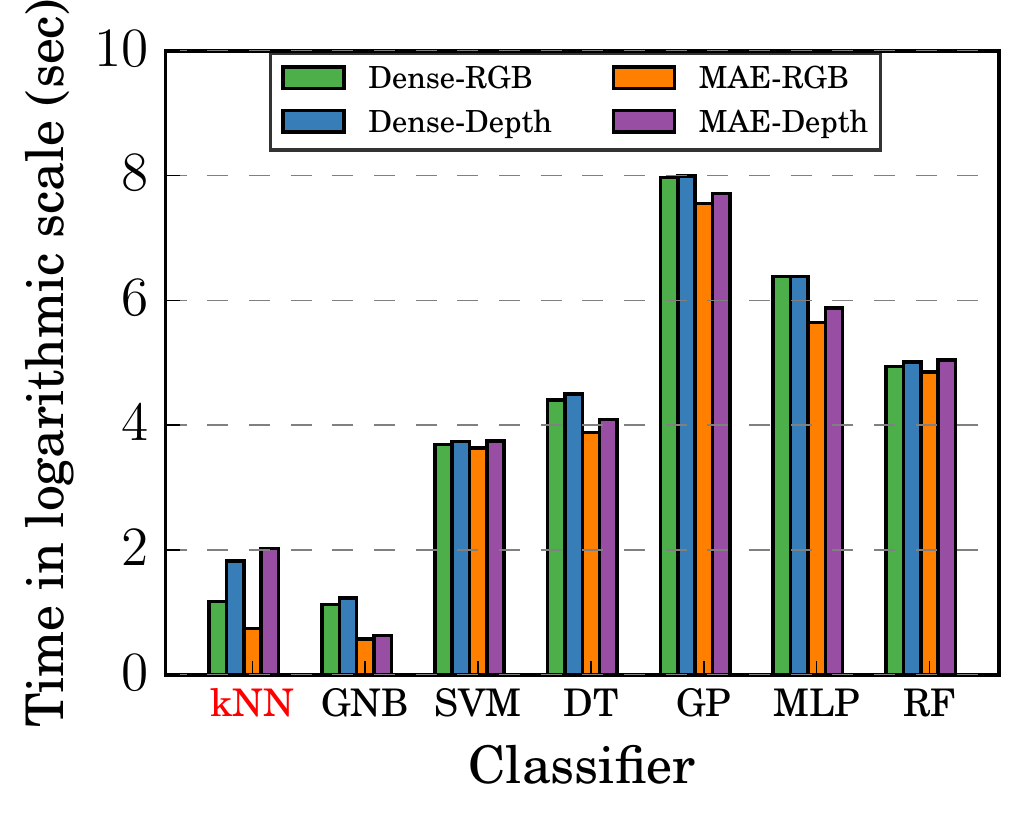} 
    \end{tabular}
    \caption{Performance of various classifiers on fine-grained recognition on restaurant object dataset: (\textit{left}) recognition accuracy; (\textit{right}) computation time. }
    \label{multi_clfs}
    \vspace{-4mm}
\end{figure}

In the first round of experiments, we encode RGB and depth modalities of the object separately using ViT (MAE) and CNN (DenseNet) and assessed the performance of seven classifiers, including k-Nearest Neighbors (kNN)~\cite{Result01}, Multi-layer Perceptron (MLP)~\cite{Result05}, Support Vector Machine (SVM)~\cite{Result02}, Decision Tree (DT)~\cite{Result03}, Gaussian Process (GP)~\cite{Result04}, Random Forest (RF)~\cite{Result06}, and Gaussian Naive Bayes (GNB)~\cite{Result07}, on the restaurant fine-grained object dataset. Results are summarized in Fig.~\ref{multi_clfs}. By comparing the performance of all classifiers, it is clear that  kNN classifier outperformed others in terms of accuracy, and provide a good blance between accuracy and computation time. It should be noted that we considered various k and distance functions, i.e., \textit{Gower, Motyka, Euclidean, Dice, Sorensen, Pearson, Neyman, Bahatta, 
KL divergence}, for the kNN classifier. Our experiments showed that kNN with Motyka distance function and k=1 achieved the best result. In the subsequent experiments, we use the kNN as the base classifier for fine-grained object recognition tasks.

\subsection{Experiments on synthetic fine-grained object datasets }
\begin{figure}[!b]
    \begin{tabular}{cc}
        \hspace{-0.4cm} 
        \includegraphics[width=0.51\linewidth]{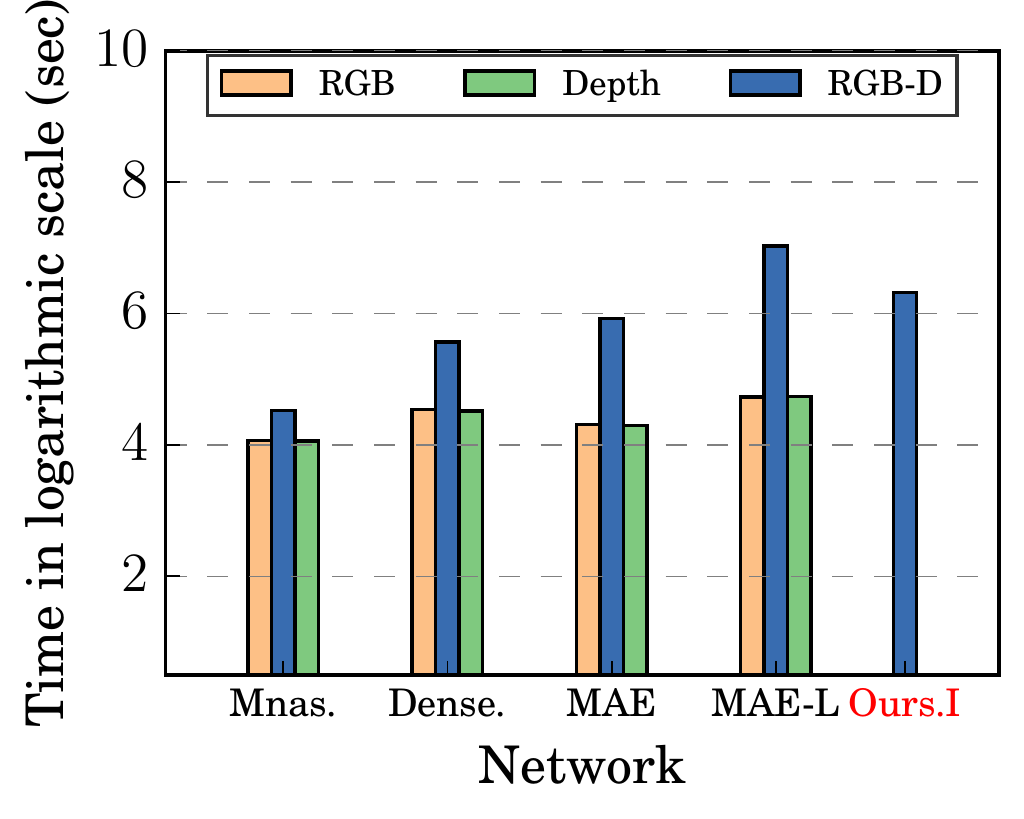} 
        \includegraphics[width=0.49\linewidth]{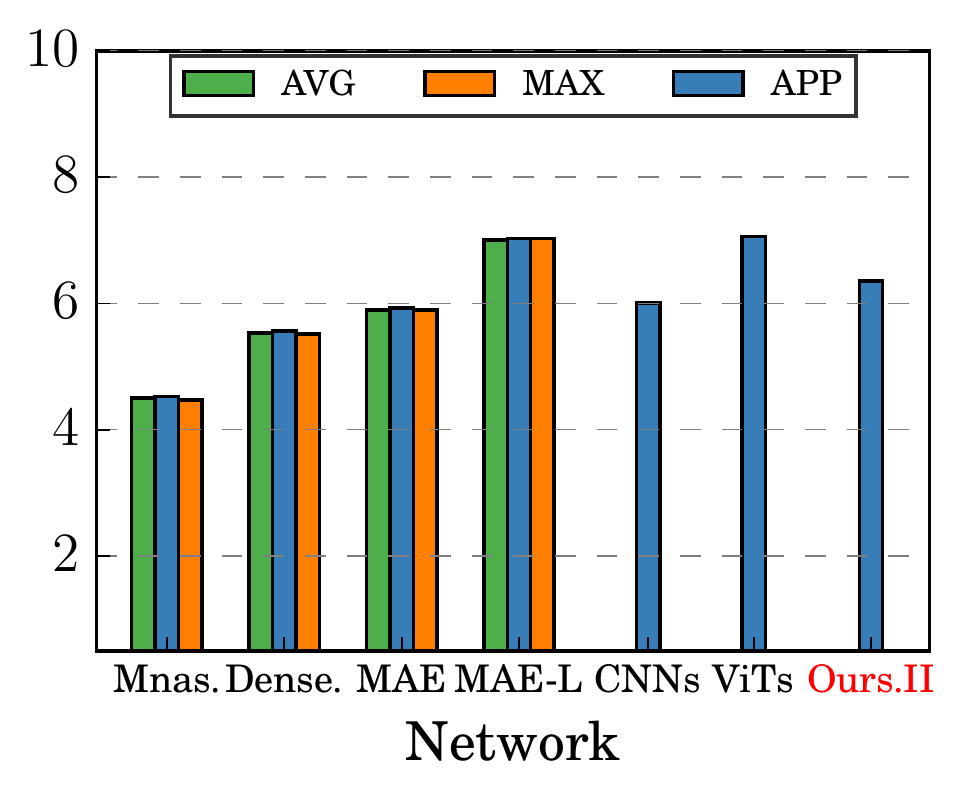}
    \end{tabular}
    \caption{ Computation time of various networks on \textbf{restaurant object dataset}: (\textit{left}) comparison of computation time for selected networks and our first approach (ours.I) based on various modalities; (\textit{right}) comparison of computation time for selected networks and our second approach (ours.II) based on different pooling functions.}
    \label{fgvc_restaurant_time}
\end{figure}


 \begin{table}[!t]
    \vspace{2mm}
    \caption{Summary of results on \textbf{Restaurant Object Dataset.}}
    \vspace{-5mm}
    \begin{center}
        \resizebox{\linewidth}{!}{
        \begin{tabular}{|c|c|c|c|c|}
        \hline
        \multicolumn{2}{|c|}{\diagbox{Networks}{Modalities}} & RGB &Depth &RGB+Depth\\
        \hline
        \multirow{2}*{CNN}& Mnas.~\cite{comp01} & 0.7968 & 0.7620 & 0.9026 \\
        
        &Dense.~\cite{comp02} &0.8842 &0.7628 &0.9151\\
        \hline
        \multirow{2}*{ViT}& MAE~\cite{comp03} & 0.8768 & 0.7495& 0.9126\\
        
        &MAE-L~\cite{comp03} &0.9159 &0.7362 &0.9176\\
        \hline
        \hline
        \multicolumn{2}{|c|}{Our approach-I \resizebox{0.6cm}{!}{\textit{\textcolor{black}{(Ours.I)}}}} & \multicolumn{2}{|c|}{-}&\textbf{\textcolor{red}{0.9342}} \\
        \hline
        \end{tabular}}
    \label{fgvc_shoe_acc_single_table}
    \end{center}
\end{table}

\begin{table}[!t]
    \vspace{-1mm}
    \caption{Summary of results on \textbf{Restaurant Object Dataset}.}
    \vspace{-5mm}
    \begin{center}
        \resizebox{\linewidth}{!}{
        \begin{tabular}{|c|c|c|c|c|}
        \hline
        \multicolumn{2}{|c|}{\diagbox{Networks}{Modalities}} & AVG &MAX &APP\\
        \hline
        \multirow{2}*{CNN}& Mnas.\cite{comp01}(\resizebox{0.7cm}{!}{\textit{RGB-D}}) & 0.8752 & 0.8835& 0.9026 \\
        
        &Dens.~\cite{comp02}(\resizebox{0.7cm}{!}{\textit{RGB-D}}) &0.8843 &0.8827 &0.9151\\
        \hline\hline
        \multicolumn{2}{|c|}{Dens.(\resizebox{0.7cm}{!}{\textit{RGB-D}})+Mnas.(\resizebox{0.7cm}{!}{\textit{RGB-D}})} & \multicolumn{2}{|c|}{-} & \textbf{0.9301} \\
        \hline
        \hline
        \multirow{2}*{ViT}& MAE~\cite{comp03}(\resizebox{0.7cm}{!}{\textit{RGB-D}}) & 0.8771 & 0.7993& 0.9126 \\
        
        &MAE-L~\cite{comp03}(\resizebox{0.7cm}{!}{\textit{RGB-D}}) &0.8769 &0.7360 &0.9176\\
        \hline
        \hline
        \multicolumn{2}{|c|}{MAE(\resizebox{0.7cm}{!}{\textit{RGB-D}})+MAE-L(\resizebox{0.7cm}{!}{\textit{RGB-D}})} & \multicolumn{2}{|c|}{-}&\textbf{0.924} \\
        \hline
        \hline
        \multicolumn{2}{|c|}{Our approach-II \resizebox{0.6cm}{!}{\textit{\textcolor{black}{(Ours.II)}}}} & \multicolumn{2}{|c|}{-}&\textbf{\textcolor{red}{0.945}} \\
        \hline
        \end{tabular}}
    \label{fgvc_shoe_acc_table}
    \end{center}
\end{table}

In first round of experiments, we assessed the performance of the proposed approaches using the fine-grained restaurant object dataset. 
Following the procedure described in previous section, we first constructed object embeddings based on ViT~(\textit{MAE}), CNN~(\textit{DenseNet}) and used kNN classifier (\textit{our first approach}). The multi-model (\textit{our first approach}) achieved 93.42$\%$ classification accuracy, which outperformed the other approaches. In the same vein, our second method with CNN~(\textit{RGB-D}) + ViT~(\textit{RGB}) obtained better classification accuracy than separate networks~(\textit{RGB-D}) and double-networks (\textit{ViT+ViT or CNN+CNN}). Additionally, as shown in Fig.~\ref{fgvc_restaurant_time}, the proposed approach also showed a good performance in terms of computation time. Especially in Fig.~\ref{fgvc_restaurant_time}(\textit{left}), our method is significantly superior to MAE-L, which has the best accuracy of all single deep networks. It can be observed that the computation time of the MAE-L approach is generally higher compared to the CNN-based approach. Moreover, the proposed approaches that uses both RGB and Depth modalities is slower than the single-modality approach. However, the multi-modal approach yields better performance than single-modal approaches in terms of accuracy. We also observe that the element-wise max pooling function is the fastest, while the average pooling function is the slowest. Overall, our proposed approaches strike a balance between computation time and accuracy.

\begin{figure*}[!t]
    \begin{tabular}{ccccc}
        \hspace{-0.35cm} \includegraphics[width=0.2\linewidth]{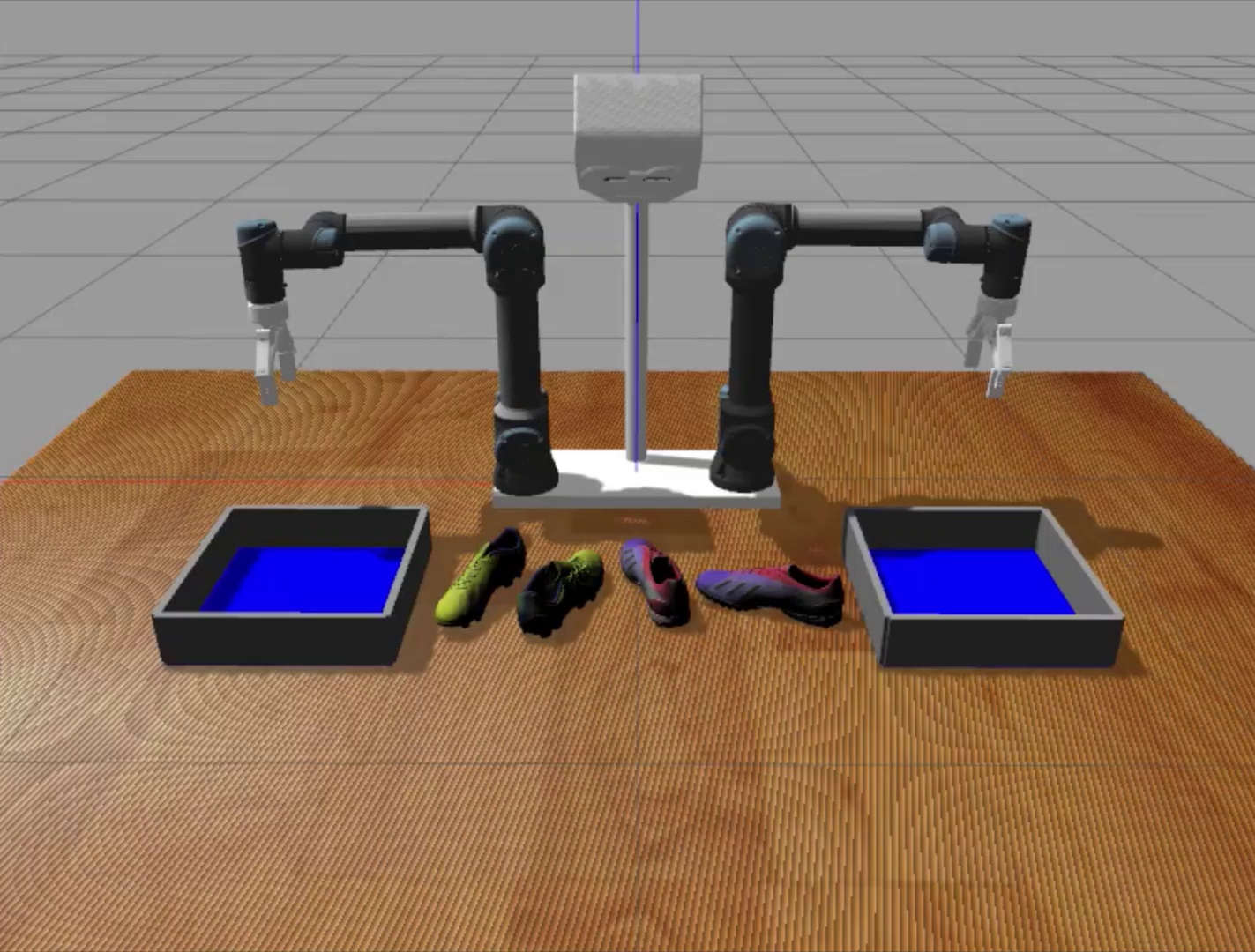}\hspace{-0.1cm}
        \includegraphics[width=0.2\linewidth, trim= 0mm 0mm 0mm 0mm, clip = true]{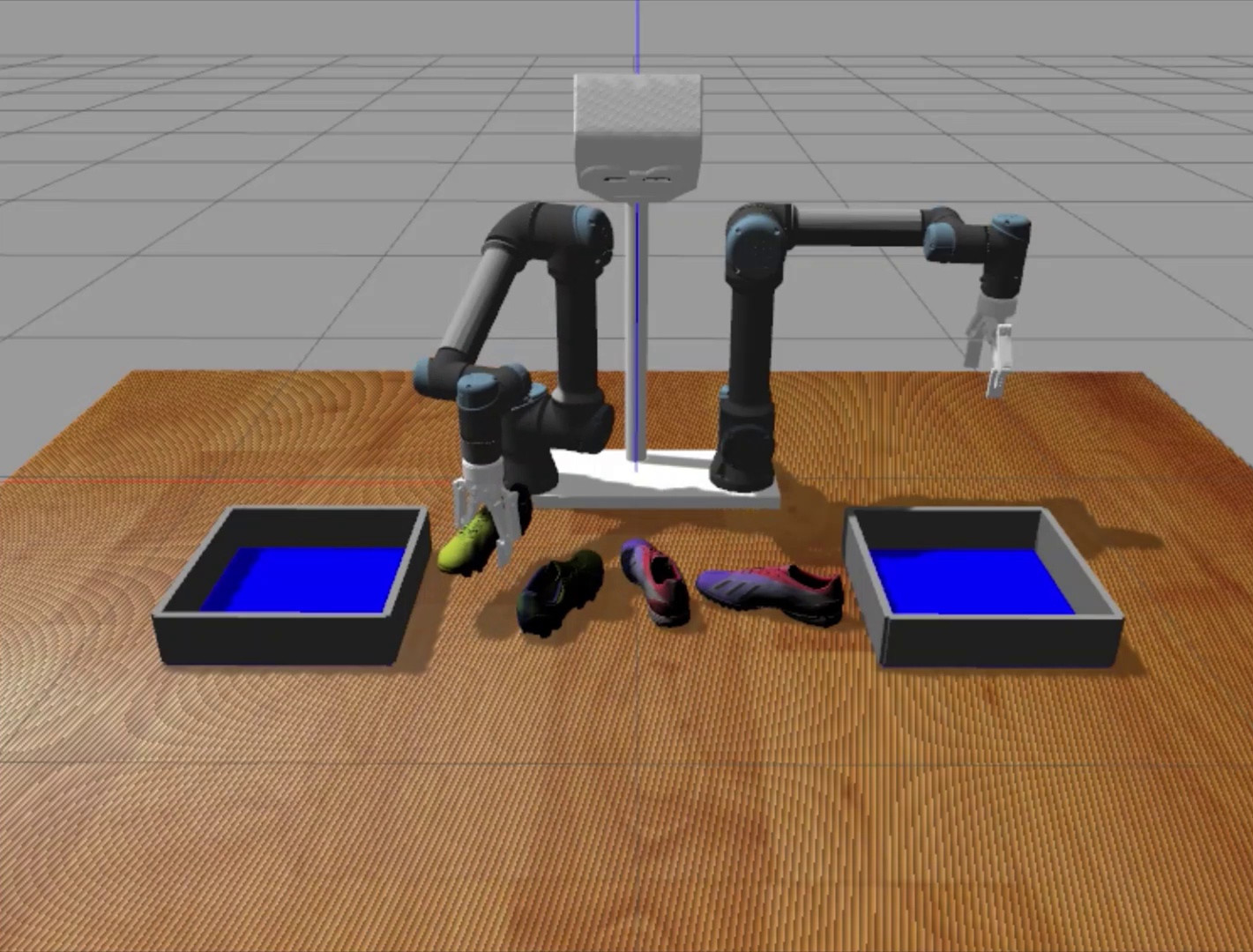}\hspace{-0.1cm}
        \includegraphics[width=0.2\linewidth, trim= 0mm 0mm 0mm 0mm, clip = true]{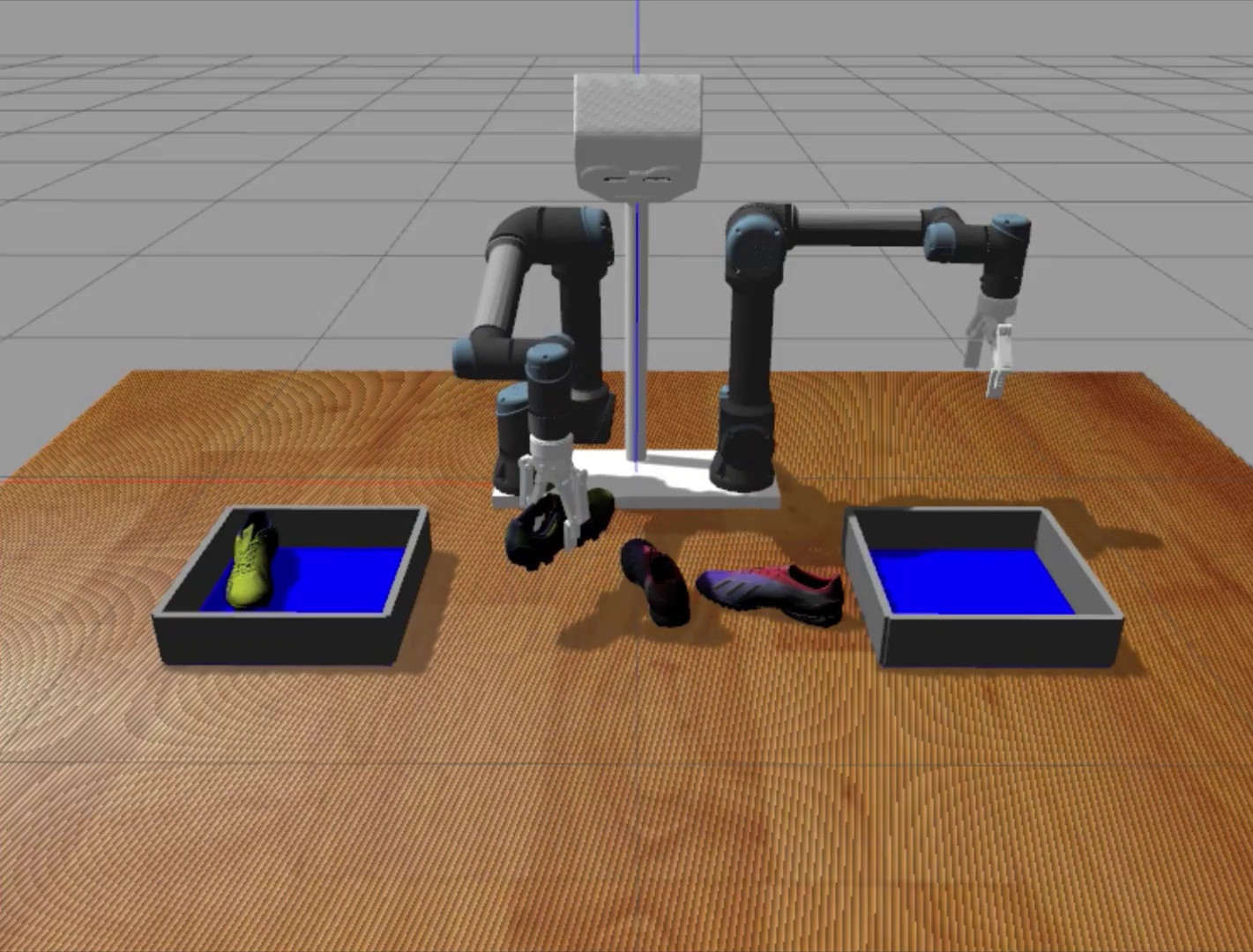}\hspace{-0.1cm}
        \includegraphics[width=0.2\linewidth]{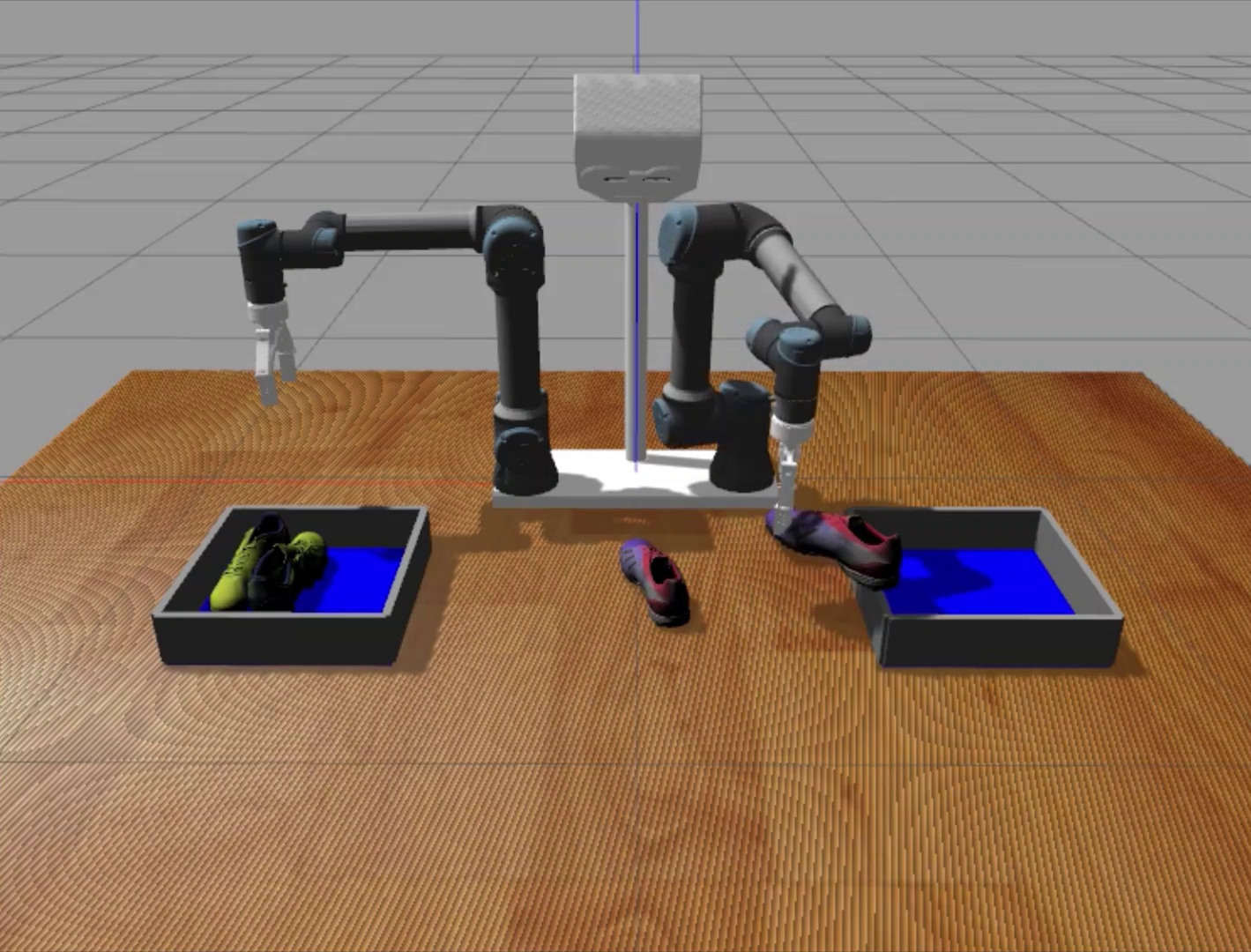}\hspace{-0.1cm}
        \includegraphics[width=0.2\linewidth, trim= 0mm 0mm 0mm 0mm, clip = true]{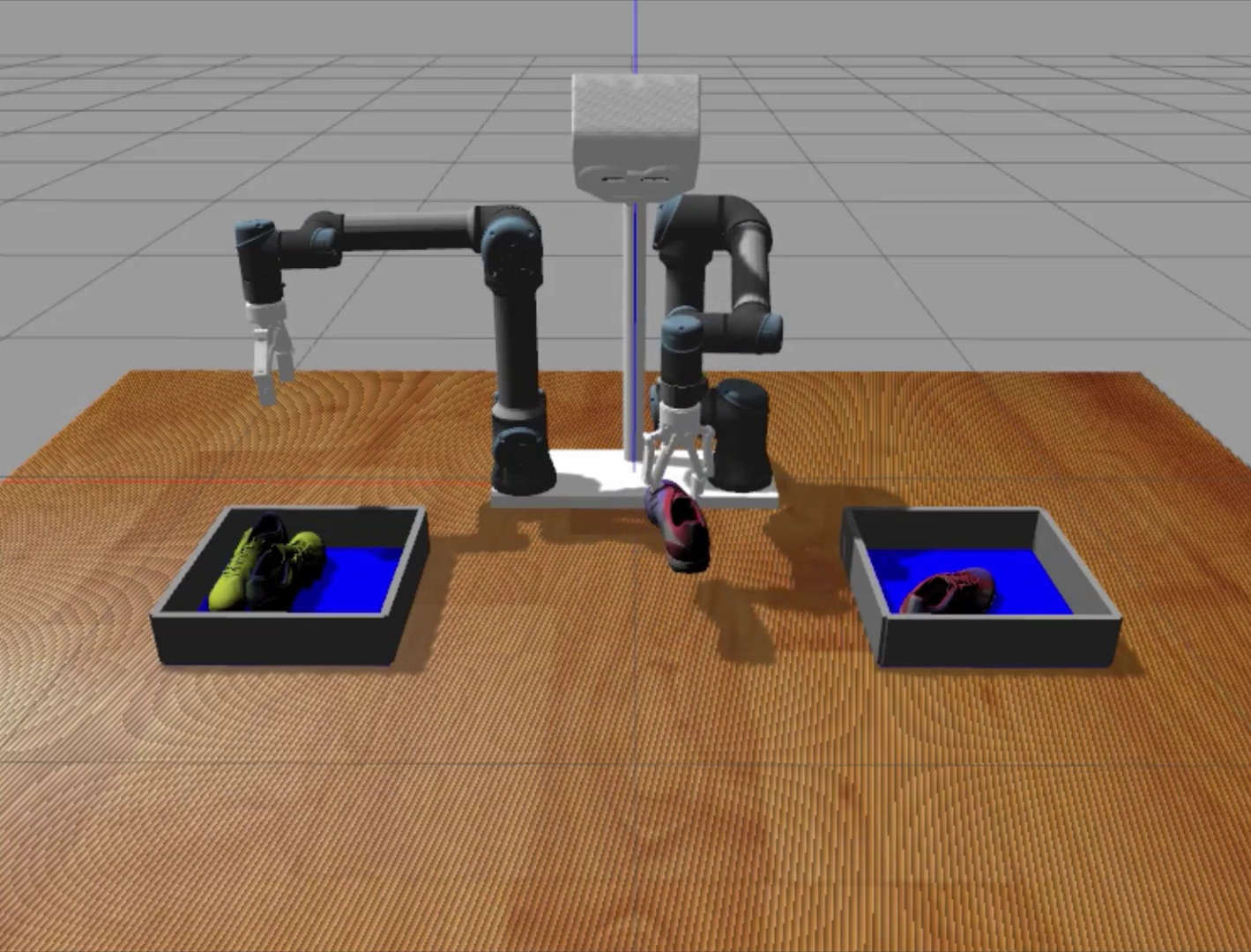}
    \end{tabular}

    \begin{tabular}{ccccc}
        \hspace{-0.35cm} \includegraphics[width=0.2\linewidth]{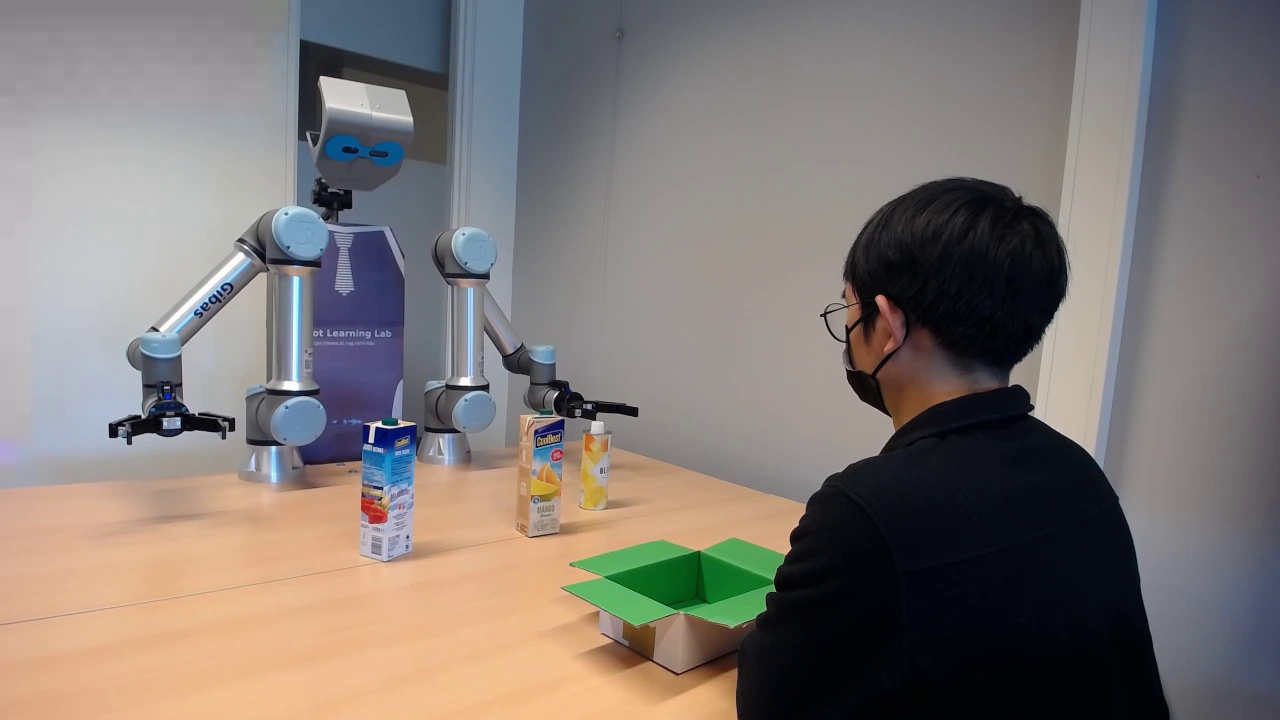}\hspace{-0.1cm}
        \includegraphics[width=0.2\linewidth, trim= 0mm 0mm 0mm 0mm, clip = true]{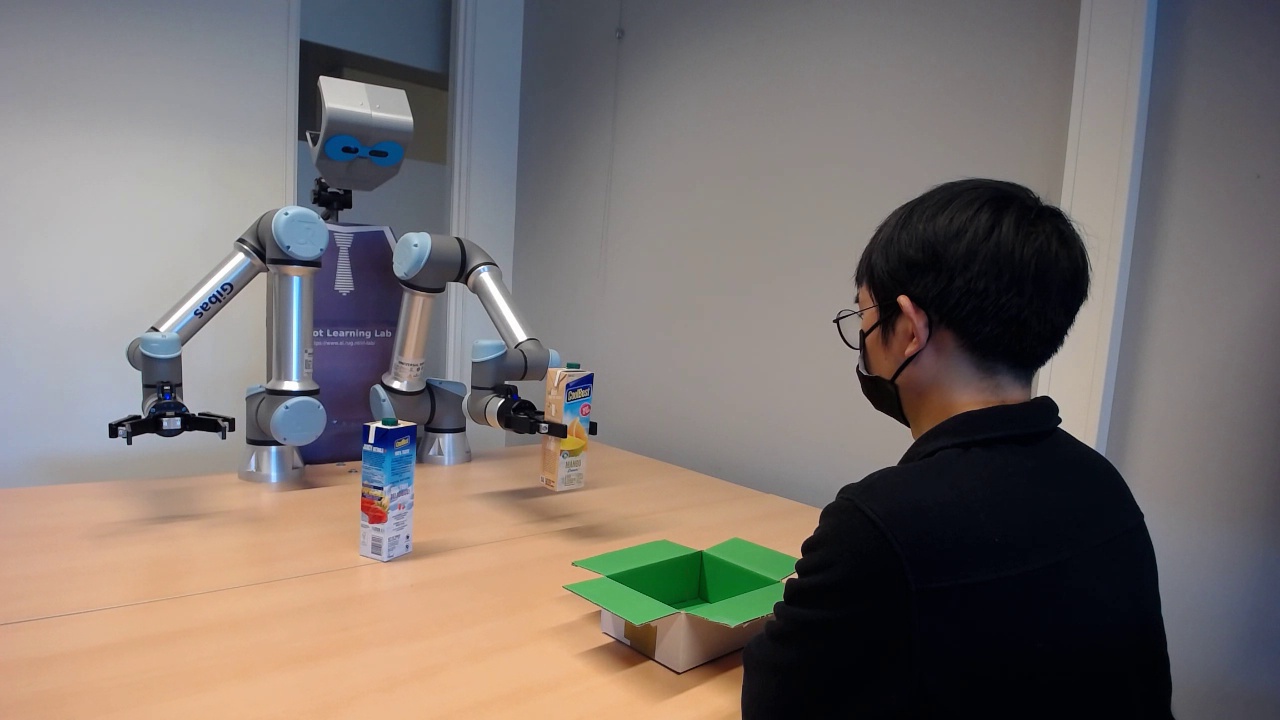}\hspace{-0.1cm}
        \includegraphics[width=0.2\linewidth]{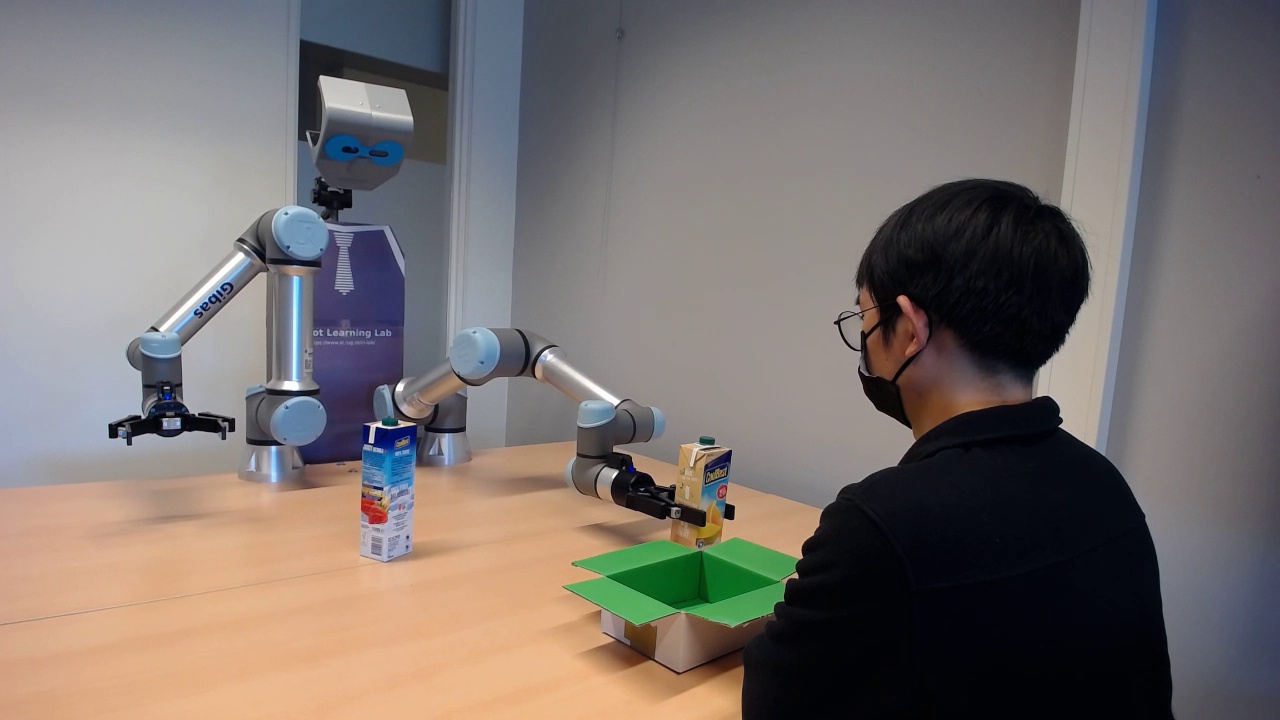}\hspace{-0.1cm}
        \includegraphics[width=0.2\linewidth, trim= 0mm 0mm 0mm 0mm, clip = true]{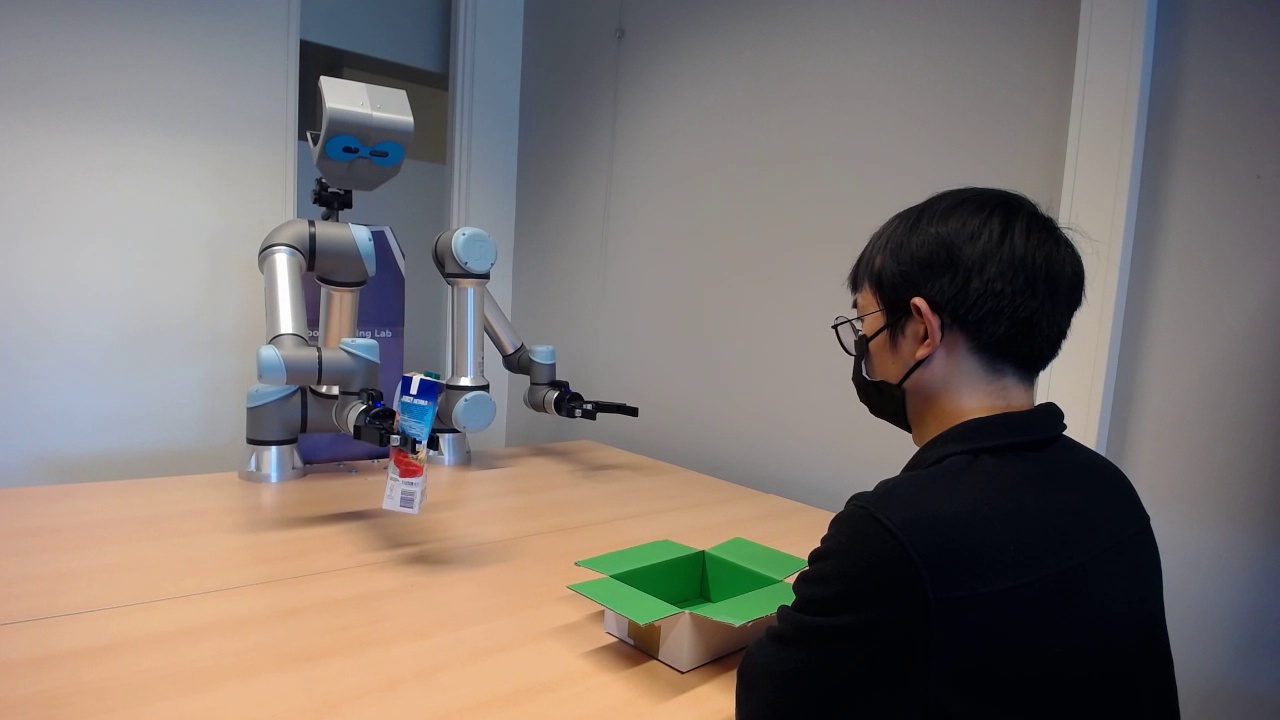}\hspace{-0.1cm}
        \includegraphics[width=0.2\linewidth]{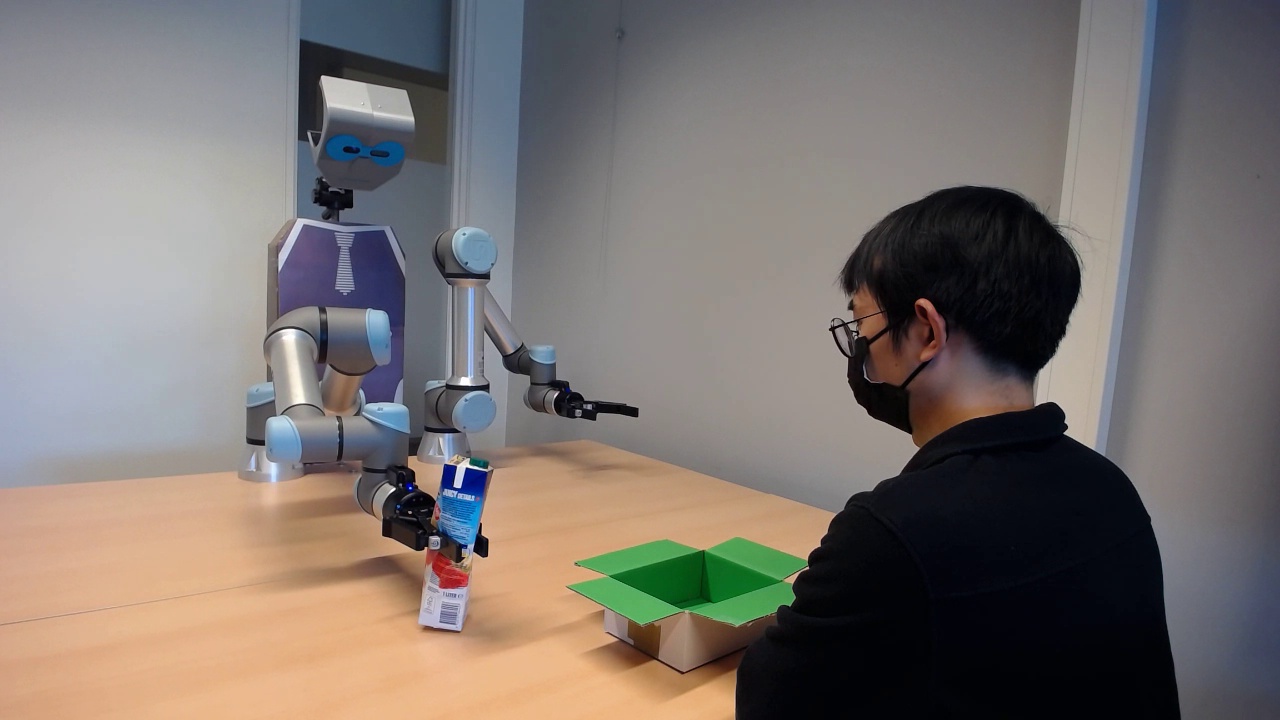}\hspace{-0.1cm}
    \end{tabular}
    \caption{A series of snapshots demonstrating the performance of our dual-arm robot in two scenarios: (\textit{top-row}) In the sorting shoes experiments, we randomly spawned four shoes in front of the robot and asked the robot to sort the shoes into the left and right baskets;
    (\textit{lower-row}) We also performed robot-assisted packaging experiments, where there are fine-grained objects (\textit{i.e. mango$\_$box and strawberry$\_$box}) in front of the robot. The objects are not reachable by the human user, and therefore, the robot should hand over the requested object to the user. 
    }
    \label{fgvc_real_robot}
\end{figure*}
\begin{figure}[!t]
    \begin{tabular}{cc}
        \hspace{-0.4cm} 
        \includegraphics[width=0.51\linewidth]{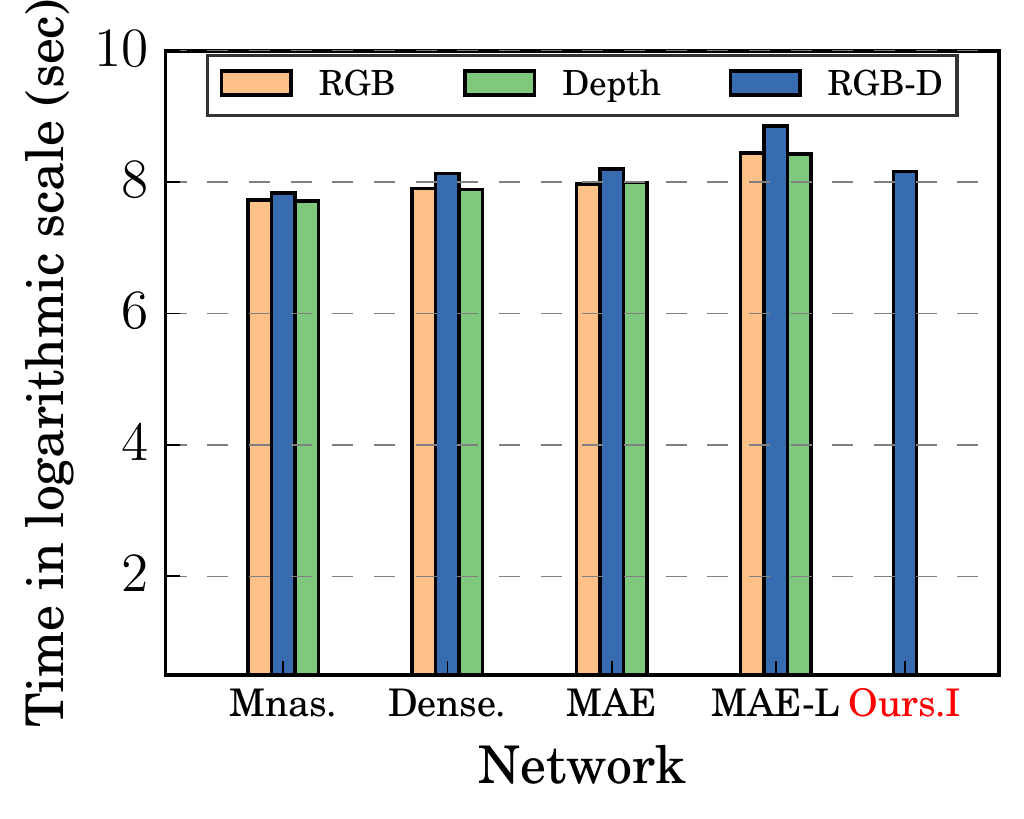} 
        \includegraphics[width=0.49\linewidth]{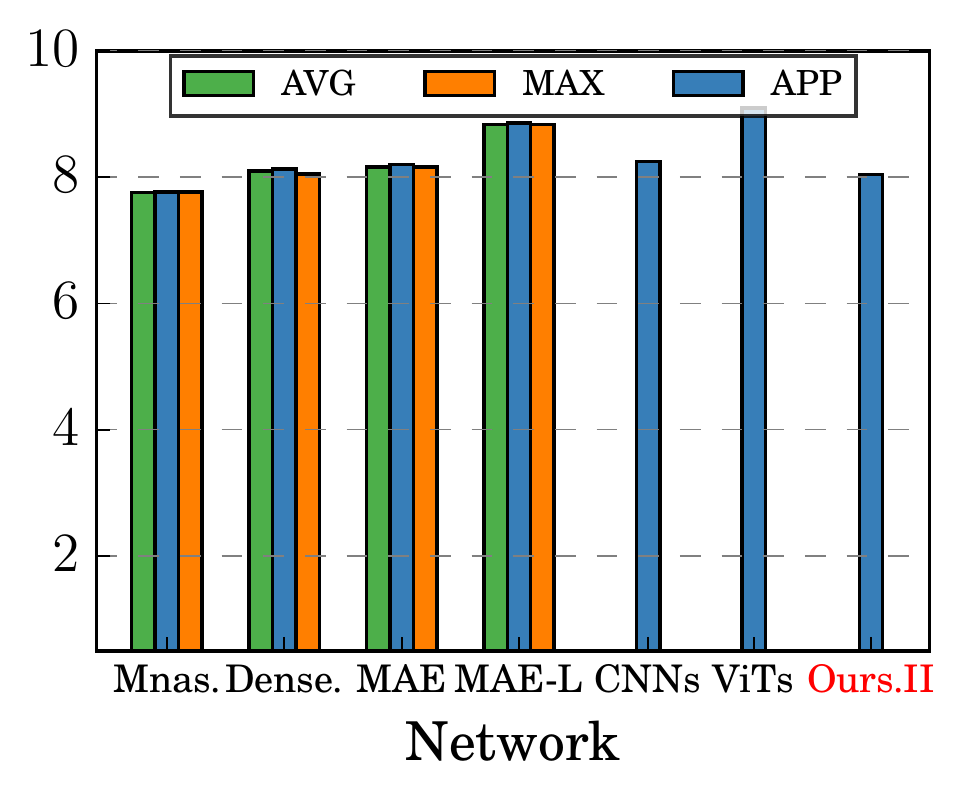} 
    \end{tabular}
    \caption{ Computation time of selected networks on \textbf{shoe object dataset}: (\textit{left}) comparison of computation time for selected networks and our first approach (ours.I) based on various modalities; (\textit{right}) comparison of computation time for selected networks and our second approach (ours.II) based on different pooling functions.}
    \label{fgvc_shoe_time}
\end{figure}

In the second round of experiments, we used the shoe object dataset to further verify the advantages of our hybrid multi-modal approach for fine-grained object categorization. In particular, we performed experiments using various networks and our hybrid approach with ViT~(\textit{RGB}) + CNN~(\textit{Depth}). 
Results are summarized on  Table~\ref{fgvc_shoe_acc_single_table}.
By comparing all the obtained results, it is clear that the proposed hybrid multi-model approach outperformed those of the single models. In particular, the accuracy of the multi-modal with ViT~(MAE) and DenseNet reached 93.13$\%$. Besides, the multi-modal have a mild increase compared to the logarithm computation time of individual models, as shown in Fig.~\ref{fgvc_shoe_time}(\textit{left}). 

 \begin{table}[!t]
    \caption{Summary of results on the \textbf{Shoe Object Dataset}.}
    \vspace{-5mm}
    \begin{center}
        \resizebox{\linewidth}{!}{
        \begin{tabular}{|c|c|c|c|c|}
        \hline
        \multicolumn{2}{|c|}{\diagbox{Networks}{Modalities}} & RGB &Depth &RGB+Depth\\
        \hline
        \multirow{2}*{CNN}& Mnas.~\cite{comp01} & 0.8226 & 0.8099 & 0.8899 \\
        
        &Dense.~\cite{comp02} &0.8564 &0.8365 &0.9031\\
        \hline
        \multirow{2}*{ViT}& MAE~\cite{comp03} & 0.9144 & 0.8069& 0.9242 \\
        
        &MAE-L~\cite{comp03} &0.9121 &0.8175 &0.9172\\
        \hline
        \hline
        \multicolumn{2}{|c|}{Our approach-I \resizebox{0.6cm}{!}{\textit{\textcolor{black}{(Ours.I)}}}} & \multicolumn{2}{|c|}{-}&\textbf{\textcolor{red}{0.9313}} \\
        \hline
        \end{tabular}}
    \label{fgvc_shoe_acc_single_table}
    \end{center}
\end{table}
\begin{table}[!t]
    \vspace{-2mm}
    \caption{Summary of results on the \textbf{Shoe Object Dataset}.}
    \vspace{-5mm}
    \begin{center}
        \resizebox{\linewidth}{!}{
        \begin{tabular}{|c|c|c|c|c|}
        \hline
        \multicolumn{2}{|c|}{\diagbox{Networks}{Modalities}} & AVG &MAX &APP\\
        \hline
        \multirow{2}*{CNN}& Mnas.~\cite{comp01}(\resizebox{0.7cm}{!}{\textit{RGB-D}}) & 0.8757 & 0.8847& 0.8899 \\
        
        &Dense.~\cite{comp02}(\resizebox{0.7cm}{!}{\textit{RGB-D}}) &0.8911 &0.899 &0.9031\\
        \hline\hline
        \multicolumn{2}{|c|}{Dense.(\resizebox{0.7cm}{!}{\textit{RGB-D}})+Mnas.(\resizebox{0.7cm}{!}{\textit{RGB-D}})} & \multicolumn{2}{|c|}{-} & \textbf{0.9021} \\
        \hline
        \hline
        \multirow{2}*{ViT}& MAE~\cite{comp03}(\resizebox{0.7cm}{!}{\textit{RGB-D}}) & 0.910 & 0.8186& 0.9242 \\
        
        &MAE-L~\cite{comp03}(\resizebox{0.7cm}{!}{\textit{RGB-D}}) &0.9042 &0.6656 &0.9172\\
        \hline
        \hline
        \multicolumn{2}{|c|}{MAE(\resizebox{0.7cm}{!}{\textit{RGB-D}})+MAE-L(\resizebox{0.7cm}{!}{\textit{RGB-D}})} & \multicolumn{2}{|c|}{-}&\textbf{0.9337} \\
        \hline
        \hline
        \multicolumn{2}{|c|}{Our approach-II \resizebox{0.6cm}{!}{\textit{\textcolor{black}{(Ours.II)}}}} & \multicolumn{2}{|c|}{-}&\textbf{\textcolor{red}{0.9351}} \\
        \hline
        \end{tabular}}
    \label{fgvc_shoe_acc_table}
    \end{center}
\end{table}
We also evaluated the performance of \textit{our approach-II}
with ViT~(\textit{RGB}) + CNN~(RBG-D). 
In this round of the experiments, all base-networks representations contained the RGB-D information. For example, RGB and Depth images, from the same view point, are input into the same CNN, or ViT networks. 
The obtained results are summarized in Table~\ref{fgvc_shoe_acc_table} and Fig.~\ref{fgvc_shoe_time}.
By comparing all results, it is clear that our multi-model outperforms the respective model method with RGB-D images. The accuracy of our multi-modal appraoch-II with DeseNet (RGB-D) and MAE (RGB) was $93.51\%$, which outperformed all single models, CNNs (\textit{Dense.+Mnas.}), and ViTs (\textit{MAE+MAE-L}). It should be noted that in terms of computation time (shown in Fig.~\ref{fgvc_shoe_time}), our methods retain superiority over ViTs with same modalities.

\subsection{Robotic demonstrations }

We  integrated our method with a dual arm robotic framework and demonstrated its potential as a fine-grained perception tool in both simulated and real-world scenarios.
Initially, we evaluated the proposed approach in the context of "\textit{sorting shoes task}". 
In this experiment, we randomly placed four fine-grained shoe objects in front of the robot as illustrated in Fig.~\ref{fgvc_overview}. We first segment the object from the scene \cite{kasaei2018towards,kasaei2020orthographicnet} and then fed them into the proposed fine-grained object recognition method. We used MAE~(RGB) + DenseNet~(RGB-D) to represent each of the objects. The CNN network is used to capture a local feature of the object and the ViT is responsible to encode the global information of the objects. The obtained representations are then used to recognize the objects. Afterward, the robot detects a proper grasp configuration for the target object, and then grasps and manipulates the object into the basket. We performed 10 round of experiments, and observed that in all experiments the robot was able to sort the shoes into the baskets successfully. A sequence of snapshots showing the performance of the robot during one of these experiments is shown in Fig.~\ref{fgvc_real_robot}(\textit{top-row}). 

We also conducted real-robot experiments aimed to package the fine-grained objects (\textit{juice boxes}) in the context of ``\textit{robot-assistant-packaging}". 
In each experiments, we randomly placed three objects, in which two of them were fine-grained juice boxes as shown in Fig.~\ref{fgvc_real_robot}(\textit{lower-row}). We provide a graphical user interface for the user to interact with the robot. The robot should recognize all objects correctly, and as soon as the user requests a specific object, it should grasp the target, and deliver it to the user. 
In this experiment, we observed that our robot could distinguish fine-grained objects form each other accurately (i.e., \textit{strawberry$\_$box} and \textit{mango$\_$box}) deliver them to the user successfully. A sequence of snapshots showing the performance of the robot during one of these experiments is shown in Fig.~\ref{fgvc_real_robot}(\textit{lower-row}). Furthermore, A video of this experiment has been attached to the paper as supplementary material.

\section{Conclusion}

In this paper, we presented a approach based on hybrid multi-modal ViT-CNN networks to handle fine-grained object classification. 
In particular, we encode the global information of the object using ViT network and encode the local representation of the object using a CNN network through both RGB and depth views of the object.  Since there is no other available fine-grained RGB-D objects dataset, we generated two synthetic fine-grained RGB-D to train and evaluate our approach. We made these datasets publicly available to the benefits of research communities. We performed several sets of experiments to evaluate the proposed approach and compare it with the single network approaches. 
We also integrated our approach in a dual-arm robotic framework and performed experiments in both simulation and real-world scenarios to show the usefulness of the proposed fine-grained 3D object recognition approach in various applications. 
Experimental results showed that our approach could distinguish fine-grained objects from each other accurately. In the continuation of this work, we would like to investigate the possibility of improving the performance by considering different views of the object from various viewpoints.

\section*{Acknowledgment}
We thank the center for Information Technology of the University of Groningen for their support and for providing access to the peregrine high performance computing cluster.

\bibliographystyle{IEEEtran}
\small\bibliography{reference}

\vspace{12pt}

\end{document}